\documentclass{article}

\usepackage{arxiv}

\usepackage{newtxtext}
\usepackage{newtxmath}

\usepackage[utf8]{inputenc} 
\usepackage[T2A,T1]{fontenc}    
\usepackage[russian,english]{babel} 
\usepackage{hyperref}       
\usepackage{url}            
\usepackage{booktabs}       
\usepackage{amsfonts}       
\usepackage{nicefrac}       
\usepackage{microtype}      
\usepackage{cleveref}       
\usepackage{lipsum}         
\usepackage{graphicx}
\usepackage{natbib}
\usepackage{doi}

\newif\ifuniqueAffiliation
\uniqueAffiliationtrue  

\title{A Systematic Benchmark of Machine Transliteration Models for the Tajik-Farsi Language Pair: A Comparative Study from Rule-Based to Transformer Architectures}

\ifuniqueAffiliation
    \author{%
        \href{https://orcid.org/0000-0003-2525-1183}{\includegraphics[scale=0.06]{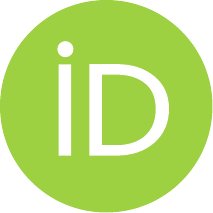}\hspace{1mm}M. K. Arabov}\thanks{Email: \texttt{MKArabov@kpfu.ru}} \\
        Institute of Computational Mathematics and Information Technologies\\
        Kazan Federal University\\
        Kazan, Russia \\
        \texttt{MKArabov@kpfu.ru}
    }
\else
    \usepackage{authblk}
    
    \newbox{\orcid}\sbox{\orcid}{\includegraphics[scale=0.06]{orcid.pdf}}
    \author[1]{%
        \href{https://orcid.org/0000-0003-2525-1183}{\usebox{\orcid}\hspace{1mm}M. K. Arabov\thanks{\texttt{MKArabov@kpfu.ru}}}%
    }
    \affil[1]{Institute of Computational Mathematics and Information Technologies, Kazan Federal University, Kazan, Russia}
\fi

\renewcommand{\shorttitle}{Transliteration Benchmark: Tajik-Farsi}

\hypersetup{
    pdftitle={A Systematic Benchmark of Machine Transliteration Models for the Tajik-Farsi Language Pair},
    pdfsubject={cs.CL},
    pdfauthor={M. K. Arabov},
    pdfkeywords={Transliteration, Tajik language, Persian language, Farsi, machine learning, natural language processing, NLP, Transformer architecture, ByT5, G2P Transformer, LSTM, mBART, comparative study, benchmarking, Cyrillic, Arabic script, parallel corpus},
}

\begin{document}
\maketitle

\begin{abstract}
This paper presents the results of the first comprehensive comparative analysis of modern machine learning architectures for the task of transliteration between the Tajik (Cyrillic script) and Persian (Arabic script) languages. The relevance of the study stems from the acute shortage of high-quality script conversion tools necessary for the development of NLP applications in the Central Asian region.

A key contribution of this work is the creation and validation of a unique parallel corpus aggregated from multiple heterogeneous sources. The corpus includes: data from crowdsourced transliteration projects, lexicographic pairs from NACIL\_ParsTranslate, parallel texts of the epic ``Shahnameh'' (Tajiki-Shahnameh), diplomatic and news articles from the website of the Ministry of Foreign Affairs of Tajikistan, texts of ``Masnavi-i Ma'navi'' from the ganjoor.net repository, official terminology lists from the Committee on Language and Terminology, and a database of transliterated correspondences from tajpers.narod.ru. The initial dataset comprised 328,253 sentence pairs. Due to computational resource limitations and the necessity of conducting multiple training cycles for various architectures, a representative subset of 40,000 pairs was formed from the general population using stratified random sampling (random\_state=42). The resulting dataset is characterized by high genre diversity: 47.1\% of the data consists of poetic fragments, 12.0\% are texts from the ``Masnavi'', 7.4\% are from the ``Shahnameh'', and the remaining portion includes prose texts, proper names, and news summaries. The average string length in Tajik was 33.4 ± 35.7 characters, and in Farsi, 28.0 ± 30.6 characters.

The experiment compared six fundamentally different classes of models: (1) a deterministic character substitution method (Rule-based Baseline); (2) a recurrent LSTM network with an attention mechanism; (3) a compact character-level Transformer (CharTransformer); (4) a G2P Transformer model proposed in this work (trained from scratch); (5) large pre-trained multilingual models (mBART, mT5 with LoRA adapters); and (6) a byte-oriented model, ByT5.

The test results demonstrate the overwhelming superiority of the ByT5 model, which achieved a chrF++ score of 87.4 ± 0.1 for the Tajik → Farsi direction and 80.1 ± 0.2 for the reverse direction. An important scientific finding is that the author's G2P Transformer model, trained exclusively on the limited 40,000-pair subset, statistically significantly outperformed the large multilingual mBART model in the forward task (72.3 vs. 62.2 chrF++). Furthermore, the complete failure of models relying on subword tokenization (mT5) was established, with scores not exceeding 18.5 chrF++, which is attributed to a fundamental mismatch between the granularity of the data and the model's tokenization strategy.

The findings convincingly demonstrate that for accurate transliteration of the Tajik--Farsi language pair, architectures operating at the byte or individual character level are unequivocally more effective than traditional multilingual Seq2Seq models that rely on subword tokenization.
\end{abstract}


\section{Introduction}
\label{sec:introduction}

The Persian language presents a unique case of digraphia, wherein two distinct writing systems coexist within a single linguistic continuum. In Iran and Afghanistan, the Perso-Arabic script is employed, whereas in Tajikistan, following the linguistic reforms of the 20th century, a modified Cyrillic alphabet has become established. This division creates a paradoxical situation: speakers of mutually intelligible dialects are deprived of access to each other's written content. The problem is particularly acute in the digital environment, where the overwhelming majority of Persian-language texts are published in Arabic script, effectively cutting off Tajik-speaking users from a significant portion of the internet space~\citep{megerdoomian-parvaz-2008-low}.

Transliteration between Tajik Cyrillic and Persian Arabic script is associated with a number of linguistic challenges. When converting from Arabic script to Cyrillic, it is necessary to restore short vowels, which are usually omitted in Perso-Arabic writing. The reverse direction is complicated by the fact that Arabic letters change shape depending on their position within a word. Furthermore, the differences extend beyond orthography to encompass certain lexical and morphological features that have developed under the influence of distinct cultural and linguistic traditions.

Despite its obvious practical significance, the task of automatic transliteration for the Tajik--Farsi language pair has long remained on the periphery of Natural Language Processing (NLP) research. The community's primary efforts have been focused on the development of machine translation tools, while the creation of transliteration systems---which could serve as a simpler and more reliable bridge between the two scripts---has not received adequate attention~\citep{merchant-tang-2024-parstext}. In recent years, interest in developing NLP tools for the Persian language has grown noticeably, as evidenced by the emergence of specialized toolkits such as DadmaTools V2~\citep{jafari-etal-2025-dadmatools}, which provides a wide range of functions for text normalization, sentiment analysis, and speech register conversion. However, none of these tools directly address the task of transliteration between Cyrillic and Arabic scripts.

A key obstacle to the development of this field is the acute shortage of high-quality parallel corpora containing texts in both graphic systems simultaneously. For a long time, researchers were forced to rely on fragmented and small-scale datasets, which precluded comprehensive training and objective comparison of modern neural network architectures. In recent years, progress has been made: resources such as ParsText~\citep{merchant-tang-2024-parstext} and a corpus based on the poem \textit{Shahnameh} have been published, although their volume and genre diversity remain limited. The relevance of the task is further confirmed by the emergence of new multi-faceted benchmarks, such as APARSIN~\citep{jafari-etal-2026-aparsin}, which includes translation and transliteration tasks for Iranian languages, thereby demonstrating the demand for systematic model comparison within this language family.

The present study aims to fill the existing gaps by conducting a systematic benchmark of a wide range of transliteration models on a unified, multi-domain dataset. For the first time, approaches from different eras are compared in a single study: from the classical rule-based character substitution method to modern architectures, such as the byte-level ByT5 and the authors' G2P Transformer model. The results of this study provide scientifically grounded recommendations for selecting an architecture for practical applications and identify the most promising directions for future development.

\section{Related Work}
\label{sec:related-work}

The first systematic approaches to the task of transliteration between Tajik Cyrillic and Persian Arabic script were proposed in the early 2010s. Davis (2012) introduced a system based on Statistical Machine Translation (SMT), treating transliteration as a character-level translation task~\citep{davis-2012-tajik}. This work also addressed the important aspect of sharing linguistic resources between the two varieties of Persian, which is particularly relevant given the significant disparity in available tools and corpora. Around the same period, the theoretical foundations for automated conversion of graphic writing systems were formulated, proposing the use of an intermediate Latin transcription; however, this approach had obvious drawbacks related to information loss and error accumulation~\citep{grashchenko2003}.

A qualitative shift in this field occurred with the proliferation of neural network methods. Merchant and Tang (2025) presented a Transformer-based Grapheme-to-Phoneme (G2P) approach, achieving chrF++ scores of 58.70 for the Farsi $\rightarrow$ Tajik direction and 74.20 for the reverse direction. The authors not only established comparable baseline metrics for future work but also analyzed in detail the differences between the two writing systems, emphasizing the non-trivial complexity of the task in both directions.

In parallel, work was underway to create specialized datasets. SadraeiJavaheri et al. (2024) proposed an original approach to addressing the data scarcity problem by creating a parallel corpus based on Ferdowsi's epic poem \textit{Shahnameh}. Using optical character recognition, verses in Tajik were extracted and then heuristically aligned with their Iranian Persian counterparts. In the same year, Merchant and Tang introduced the ParsText corpus, containing 2,813 Persian sentences written in both graphic systems and manually collected from blog platforms and news websites~\citep{merchant-tang-2024-parstext}. The authors rightly noted that printed digraphic texts in Tajikistan are far more common in physical form than in digital form, and their work was intended to at least partially fill this gap.

The most recent work at the time of writing this article is ``ParsTranslit'' by Merchant and Tang (2026), which presents a modern sequence-to-sequence model trained on all available datasets, as well as two new proprietary datasets~\citep{merchant-tang-2026-parstranslit}. A critically important contribution of this work is the demonstration that models trained on narrow domains lack the universality required for real-world use, and only the aggregation of data from various sources enables the creation of a truly universal transliteration system. The presented model achieved impressive scores: chrF++ of 87.91 and normalized CER of 0.05 for the Farsi $\rightarrow$ Tajik direction, and 92.28 and 0.04 for the reverse direction.

An important contribution to the development of the resource base and evaluation methodology for the Tajik--Farsi pair has been made by previous works of the authors of the present study. In the work ``TajPersLexon'' (2026), the authors presented a curated Tajik--Persian parallel lexical resource comprising 40,112 word and short phrase pairs, intended for cross-script lexical search, transliteration, and alignment in low-resource settings~\citep{arabov-2026-tajperslexon}. As part of this work, a comprehensive CPU-only benchmark was conducted, comparing three methodological families: a lightweight hybrid pipeline, neural sequence-to-sequence models, and information retrieval methods. In another work by the authors (Kurbonovich, 2026), transliteration from Tajik to Persian was investigated using a character-level Transformer trained on a corpus of 52,152 lexicographically verified pairs~\citep{kurbonovich-2026-character}. The best model configuration with beam search ($k=3$) achieved a CER of 0.3182 and an exact match accuracy of 0.3215, outperforming both dictionary-based rule-based methods and recurrent neural baselines.

In addition to works directly focused on transliteration, it is worth noting the existence of several tools for Persian text processing that can be used in pre- and post-processing stages. The DadmaTools V2 toolkit deserves special attention due to its use of adapter techniques (LoRA), which significantly reduce memory consumption by employing a shared pre-trained model with the addition of lightweight adapter layers for specific tasks~\citep{jafari-etal-2025-dadmatools, hu2022_lora}. DadmaTools V2 includes modules for sentiment analysis, conversion of informal text to formal, and spell checking, making it a powerful and versatile resource for the advancement of Persian NLP. Although these tools do not directly address the task of transliteration between Cyrillic and Arabic scripts, they form an important part of the broader Persian NLP ecosystem.

Among recent works confirming the relevance of the task, the APARSIN benchmark should be highlighted~\citep{jafari-etal-2026-aparsin}. This multi-faceted dataset for Iranian languages includes tasks for machine translation, transliteration, and sentiment analysis, reflecting the growing interest of the research community in systematic model comparison for this language family. The emergence of such resources further motivates the conduct of the present study.

Finally, it is important to situate the present work within the broader context of computational studies of Persian language contact. Basirat et al.~\citep{basirat-etal-2026-computational} recently proposed a computational framework for quantifying structural traces of language contact in monolingual Persian language models. Their findings that morphological features such as case and gender are strongly shaped by language-specific structure, whereas universal syntax remains largely insensitive to historical contact, underscore the linguistic complexity inherent in processing Persian varieties. This further motivates the need for specialized models capable of handling the orthographic and morphological divergences between Tajik and Farsi.

An analysis of the existing literature shows that, despite significant progress in recent years, several fundamental problems persist in the field. Most works either focus on narrow domains or do not conduct a systematic comparison of architectures under controlled conditions on a single multi-domain dataset. Reproducibility of results also remains an issue: many studies do not publish code and precise experimental configurations. The present work aims to fill these gaps by conducting the first systematic benchmarking of a wide range of transliteration models for the Tajik--Farsi pair with full disclosure of methodology and data.

\section{Data}
\label{sec:data}

To conduct an objective benchmark of transliteration models for the Tajik--Farsi language pair, the availability of a representative parallel corpus spanning various functional styles and genres is critically important. In the framework of the present study, a unique dataset was aggregated, combining texts from a wide range of sources. It includes official and news materials from the website of the Ministry of Foreign Affairs of the Republic of Tajikistan, terminology lists from the Committee on Language and Terminology under the Government of the Republic of Tajikistan, fragments of classical Persian poetry---the epic \textit{Shahnameh} by Ferdowsi, \textit{Masnavi-i Ma'navi} by Jalal ad-Din Rumi (from the ganjoor.net repository), and selected rubaiyat of Omar Khayyam---as well as data from open crowdsourcing and lexicographic projects, and the parallel lexical corpus \texttt{TajPersParallelLexicalCorpus}, previously published on the Hugging Face platform~\citep{arabov2026tajpersparallelcorpus}. This effort builds upon our prior experience in constructing large-scale text resources for Tajik, including a multiformat corpus designed specifically for training modern language models~\citep{arabov-2025-multiformat}.

All collected texts underwent a multi-stage cleaning and normalization procedure. Incorrect characters and formatting artifacts were removed, various types of spaces (including non-breaking and narrow spaces characteristic of Perso-Arabic script) were unified, and strings containing fragments in Latin script or characters from other alphabets were excluded. After initial filtering and deduplication, the total size of the corpus amounted to 328,253 sentence pairs. Due to the technical limitations of the computational resources available for a full-scale experiment (including the need for multiple training runs of over a dozen models with several random initializations), using the entire data volume was deemed impractical. To ensure the reproducibility and controllability of the experiment, a representative subset of 40,000 pairs was formed from the general population using pseudo-random sampling with a fixed seed (\texttt{random\_state=42}). As will be shown below, this subsample fully preserves the statistical characteristics of the original array, making it legitimate to extend the conclusions drawn from it to the entire corpus.

To confirm the representativeness of the formed subsample, a comparative analysis of the main statistical indicators of the full dataset (328,253 pairs) and the subsample (40,000 pairs) was conducted. The comparison results are presented in Tables~\ref{tab:domain-distribution} and~\ref{tab:length-statistics}.

\begin{table}[htbp]
\centering
\caption{Comparison of record distribution by categories (domains) in the full dataset and the subsample}
\label{tab:domain-distribution}
\begin{tabular}{lcc}
\toprule
\textbf{Category} & \textbf{Full dataset (328,253)} & \textbf{Subsample (40,000)} \\
\midrule
Poetic fragments (\texttt{poetry\_parts}) & 47.2\% & 47.1\% \\
Rumi's \textit{Masnavi} (\texttt{masnavi}) & 11.9\% & 12.0\% \\
Unique Tajik words (\texttt{unique\_tajik\_words}) & 11.6\% & 11.5\% \\
\textit{Shahnameh} (\texttt{shahnameh}) & 7.6\% & 7.4\% \\
Prose fragments (\texttt{prose\_parts}) & 7.3\% & 7.4\% \\
Proper names---Persons (\texttt{paranames\_per}) & 6.1\% & 6.2\% \\
General vocabulary (\texttt{words}) & 4.4\% & 4.3\% \\
Proper names---Locations (\texttt{paranames\_loc}) & 2.9\% & 3.0\% \\
Other categories (\texttt{dr}, \texttt{jj}, \texttt{paranames\_org}, \texttt{bbc}) & 1.0\% & 1.1\% \\
\bottomrule
\end{tabular}
\end{table}

As seen from Table~\ref{tab:domain-distribution}, the subsample reproduces the genre structure of the original corpus with high accuracy---the maximum deviation in the share of individual categories does not exceed 0.2 percentage points.

\begin{table}[htbp]
\centering
\caption{Comparison of descriptive statistics of character sequence lengths}
\label{tab:length-statistics}
\begin{tabular}{lcccc}
\toprule
\textbf{Statistic} & \textbf{Tajik (328k)} & \textbf{Tajik (40k)} & \textbf{Farsi (328k)} & \textbf{Farsi (40k)} \\
\midrule
Mean & 33.35 & 33.38 & 27.98 & 28.00 \\
Standard deviation & 35.95 & 35.71 & 30.84 & 30.58 \\
Median & 26 & 26 & 22 & 22 \\
25th percentile & 19 & 19 & 17 & 17 \\
75th percentile & 34 & 34 & 28 & 28 \\
Maximum length & 973 & 755 & 825 & 652 \\
\bottomrule
\end{tabular}
\end{table}

The data in Table~\ref{tab:length-statistics} demonstrate that the distribution of string lengths in the subsample is virtually identical to the full dataset. The slight decrease in maximum values in the subsample is explained by the exclusion of a small number of anomalously long sequences, which is standard practice in data preparation for training neural network models and does not affect representativeness.

Since all experiments within the framework of this work were conducted specifically on the subsample of 40,000 pairs, its detailed descriptive statistics are provided below. The total number of pairs is 40,000, of which there are 40,000 unique Tajik strings and 39,996 unique Farsi strings (four pairs have identical Farsi translations with differing Tajik originals). The average string length in Tajik was $33.38 \pm 35.71$ characters (median 26), and in Farsi---$28.00 \pm 30.58$ characters (median 22). The average number of words per string: $5.8 \pm 5.7$ for Tajik and $6.3 \pm 6.3$ for Farsi. The greater character length of Tajik strings with a comparable number of words is explained by the features of Cyrillic script, in which all vowels are written explicitly, unlike Perso-Arabic writing, where short vowels are omitted in writing.

Character frequency analysis, presented in Table~\ref{tab:char-frequency}, confirms the expected typological patterns. In the Tajik subcorpus, the vowels {\fontencoding{T2A}\selectfont ``а''}, {\fontencoding{T2A}\selectfont ``и''}, and {\fontencoding{T2A}\selectfont ``о''} dominate, while in Persian, the letter \textit{alef} and high-frequency consonants \textit{re} and \textit{ye} prevail. Analysis of script usage showed that the average number of Cyrillic characters per Tajik string is 27.44, and the number of Arabic script characters per Farsi string is 18.76.

\begin{table}[htbp]
\centering
\caption{Top 10 most frequent characters in the subsample (40,000 pairs)}
\label{tab:char-frequency}
\begin{tabular}{clccl}
\toprule
\textbf{\#} & \textbf{Tajik character} & \textbf{Frequency} & \textbf{Farsi character} & \textbf{Frequency} \\
\midrule
1 & Space & 190,188 & Space & 212,338 \\
2 & {\fontencoding{T2A}\selectfont а} & 167,061 & \textit{alef} & 113,481 \\
3 & {\fontencoding{T2A}\selectfont и} & 94,200 & \textit{re} & 78,884 \\
4 & {\fontencoding{T2A}\selectfont о} & 89,231 & \textit{ye} & 77,101 \\
5 & {\fontencoding{T2A}\selectfont р} & 78,567 & \textit{nun} & 73,665 \\
6 & {\fontencoding{T2A}\selectfont н} & 72,378 & \textit{dal} & 64,929 \\
7 & {\fontencoding{T2A}\selectfont д} & 63,014 & \textit{vav} & 56,331 \\
8 & {\fontencoding{T2A}\selectfont у} & 51,412 & \textit{he} & 53,069 \\
9 & {\fontencoding{T2A}\selectfont м} & 41,778 & \textit{be} & 44,098 \\
10 & {\fontencoding{T2A}\selectfont т} & 41,242 & \textit{mim} & 43,915 \\
\bottomrule
\end{tabular}
\end{table}

Thus, the formed dataset is balanced by genre, contains both short lexical units and full-fledged sentences, and reflects the real complexity of the transliteration task, including the need to process proper names, poetic archaisms, and modern journalism. The confirmed statistical identity of the subsample to the full corpus allows us to consider the experimental results obtained on 40,000 pairs as valid for the entire general population and ensures a high degree of confidence in the conclusions drawn during the benchmark.

\section{Model Architectures and Experimental Protocol}
\label{sec:architectures}

To conduct an objective comparative analysis within a unified software framework, six classes of models were implemented or adapted, encompassing both classical approaches and modern neural network architectures. This section sequentially describes the architectural features of each model, as well as the training protocol and evaluation metrics used in the experiments.

\subsection{Rule-based Baseline}
\label{subsec:rule-based}

As a lower bound of quality, a deterministic character-by-character substitution method was implemented, based on static dictionaries of correspondences between Cyrillic and Arabic graphemes. The dictionary for the Tajik $\rightarrow$ Farsi direction includes 70 rules, covering both lowercase and uppercase letters of the Tajik alphabet, including specific graphemes ({\fontencoding{T2A}\selectfont ғ}, {\fontencoding{T2A}\selectfont ӣ}, {\fontencoding{T2A}\selectfont қ}, {\fontencoding{T2A}\selectfont ӯ}, {\fontencoding{T2A}\selectfont ҳ}, {\fontencoding{T2A}\selectfont ҷ}) and their positional variants. The reverse dictionary contains 47 rules and accounts for the most frequent Arabic characters. Transliteration is performed by sequentially iterating over the input string and replacing each character with the corresponding element from the dictionary; if no rule is found, the character is left unchanged. This approach requires no training and does not use contextual information, yet serves as an important reference point for assessing the quality gain achieved through neural network architectures.

\subsection{LSTM with Attention Mechanism}
\label{subsec:lstm}

The model based on Long Short-Term Memory implements an encoder-decoder architecture with the additive attention mechanism of Bahdanau et al.~\citep{bahdanau2015attention}. The encoder consists of a two-layer bidirectional LSTM with a hidden size of 256, receiving indices of source alphabet characters mapped to embeddings of dimension 256. The decoder also contains two LSTM layers with a hidden state size of 256. At each decoding step, a context vector is computed as a weighted sum of the encoder outputs, where the weights are determined by an attention mechanism comparing the current decoder hidden state with all encoder outputs. The context vector is concatenated with the embedding of the previous token and fed into the decoder LSTM cell. The output layer is a linear transformation of the concatenation of the LSTM output and the context vector into the vocabulary space dimension. During training, teacher forcing is applied with a probability of 0.5. Optimization is performed using AdamW with an initial learning rate of $5 \times 10^{-4}$ and gradient clipping (norm 1.0). To prevent overfitting, dropout with a probability of 0.1 is applied to all layers except the output layer.

\subsection{Character-Level Transformer (CharTransformer)}
\label{subsec:char-transformer}

This model is a compact implementation of the Transformer architecture, trained from scratch on a character vocabulary. The encoder and decoder each consist of four layers, with an embedding dimension of 256, eight attention heads, and a hidden layer dimension of 1024 in the feed-forward blocks. Positional encoding is implemented using sinusoidal functions, allowing the model to process sequences of arbitrary length (up to 5,000 characters). Regularization is achieved through dropout with a probability of 0.1. Training is conducted using the AdamW optimizer with an initial learning rate of $5 \times 10^{-4}$ and a cosine annealing scheduler. An important feature of the implementation is the masking of padding tokens in the attention mechanism, which prevents the consideration of uninformative positions during context computation.

\subsection{G2P Transformer}
\label{subsec:g2p-transformer}

The G2P Transformer model, developed within the framework of this study, is an enhanced version of the CharTransformer, specifically adapted for the task of transliteration as a special case of grapheme-to-grapheme conversion. Architecturally, the model is identical to the CharTransformer (four-layer encoder-decoder, 256-dimensional embeddings, 8 attention heads), but incorporates several important modifications. First, the maximum sequence length has been increased to 512 characters, which avoids truncation of long sentences. Second, label smoothing regularization with a parameter of 0.1 is applied, preventing the model from becoming overconfident and improving generalization ability. Third, L2 regularization (weight decay) with a coefficient of $10^{-4}$ is added to the AdamW optimizer. The initialization of embedding weights and the output layer is performed using a uniform distribution in the range $[-0.1, 0.1]$, which promotes more stable training onset. During inference, greedy decoding (argmax) is employed without beam search, which, as preliminary experiments showed, did not lead to a noticeable decrease in quality while significantly accelerating generation.

\subsection{Pretrained Multilingual Models}
\label{subsec:pretrained}

Three pretrained models of the Transformer family, available through the Hugging Face Transformers library, were employed in the experiments. \texttt{mBART-large-50-many-to-many-mmt} is a model with 610 million parameters, pretrained on 50 languages, including Tajik (code \texttt{tg\_Cyrl}) and Persian (code \texttt{fa\_IR}). For correct handling of the transliteration direction, the source and target language identifiers, as well as the forced beginning-of-sequence token, were explicitly specified in the tokenizer and generation configuration. \texttt{mT5-small} is a model with 300 million parameters, pretrained on the multilingual Common Crawl corpus. Due to its large size, full fine-tuning of all parameters was impractical; therefore, the Low-Rank Adaptation (LoRA) technique~\citep{hu2022_lora} was applied with a rank of 8, a scaling factor of 32, and a dropout of 0.05. Adapters were added to the query, key, value, and output projection matrices of the attention mechanism (modules \texttt{q}, \texttt{k}, \texttt{v}, \texttt{o} for mT5). \texttt{ByT5-small} is a model of similar size (300 million parameters) that operates directly on the UTF-8 bytes of the input text without subword tokenization. Thanks to its byte-level representation, ByT5 does not require vocabulary construction and is capable of processing arbitrary characters, including rare graphemes of the Tajik alphabet.

All pretrained models were fine-tuned using the \texttt{Seq2SeqTrainer} class for two epochs with an initial learning rate of $3 \times 10^{-4}$ and linear decay. Dynamic batching was employed with a maximum batch size of 8, and for mBART and ByT5, mixed-precision mode (bfloat16) was activated, allowing the models to fit within the available GPU memory. Checkpoints were saved at the end of each epoch, and the best model was selected based on the validation loss value.

\subsection{Experimental Protocol and Evaluation Metrics}
\label{subsec:protocol}

All experiments were conducted on a compute node with an NVIDIA GPU having at least 18~GB of video memory. To ensure reproducibility of results, for each model and each transliteration direction, training and evaluation were performed on three independent random initializations with fixed pseudo-random number generator seeds: 42, 43, and 44. On each initialization, a stratified split of the original 40,000-pair sample into training (80\%), validation (10\%), and test (10\%) subsets was performed, preserving category proportions. Intermediate results were serialized in JSON format, guaranteeing exact preservation of numerical values and allowing repeated computation to be avoided when restarting experiments.

The quality of transliteration was assessed on the test set using six metrics: chrF++ (weighted average of character-level and word-level precision and recall), BLEU, TER, CER (Character Error Rate), WER (Word Error Rate), and exact match accuracy. For computing chrF++, BLEU, and TER, the SacreBLEU library was used with default parameters; for CER and WER, Levenshtein distance normalized by the reference length was employed. For each model and direction, the mean values and standard deviations of the metrics were computed over three runs, as well as 95\% confidence intervals using the bootstrap method with 2,000 resamples. The statistical significance of differences between models was assessed using paired Wilcoxon signed-rank tests and paired t-tests for the chrF++ metric, with a significance level of 0.05.

The primary metric selected was chrF++, as it accounts for both character-level and word-level n-gram overlap, which is critical for character-by-character conversion tasks where morphological regularity and suffixal patterns play a key role~\citep{popovic2017chrf}.

\section{Experimental Results}
\label{sec:results}

This section presents the results of the empirical evaluation of six classes of transliteration models compared within a unified computational framework. In accordance with the protocol outlined in Section~\ref{subsec:protocol}, all architectures were tested on stratified test subsets (10\% of 40,000 pairs) with three repetitions for each conversion direction (Tajik $\rightarrow$ Farsi and Farsi $\rightarrow$ Tajik). The final metrics were averaged over independent runs with different random generator seeds (42, 43, 44), and for the primary metric chrF++, 95\% confidence intervals were calculated using the bootstrap method (2,000 iterations). Additionally, temporal and resource costs were recorded: training duration, inference latency on a single example, and peak GPU memory consumption.

To ensure compact presentation of data in a two-column format, the results are grouped into three thematic tables: primary quality metrics (Table~\ref{tab:primary-metrics}), auxiliary error metrics (Table~\ref{tab:aux-metrics}), and computational characteristics (Table~\ref{tab:compute-metrics}). Each block is accompanied by a detailed analysis and comparison of architectures.

\subsection{Primary Transliteration Quality Metrics}
\label{subsec:primary-metrics}

Table~\ref{tab:primary-metrics} presents the values of key metrics that directly characterize text conversion accuracy: chrF++, BLEU, CER, and Acc\%. The chrF++ metric serves as the primary one, as it accounts for both character-level and word-level n-gram overlap, which is especially important for character-by-character conversion tasks with pronounced morphological regularity~\citep{popovic2017chrf}. BLEU is traditionally used in machine translation and provides insight into the overlap of longer sequences, while CER reflects the proportion of erroneously transliterated characters, and Acc\% represents the strict proportion of perfectly matching strings.

\begin{table}[htbp]
\centering
\caption{Primary transliteration quality metrics (mean $\pm$ std. dev. over three runs; 95\% confidence intervals for chrF++ in square brackets).}
\label{tab:primary-metrics}
\begin{tabular}{lccccc}
\toprule
\textbf{Model} & \textbf{Dir.} & \textbf{chrF++} & \textbf{BLEU} & \textbf{CER} & \textbf{Acc\%} \\
\midrule
Baseline & f$\rightarrow$t & 10.4 $\pm$ 2.4 [8.6--13.8] & 0.36 $\pm$ 0.09 & 0.54 $\pm$ 0.08 & 0.04 $\pm$ 0.01 \\
Baseline & t$\rightarrow$f & 14.0 $\pm$ 6.0 [9.7--22.5] & 0.42 $\pm$ 0.19 & 0.62 $\pm$ 0.14 & 0.46 $\pm$ 0.42 \\
CharTransformer & f$\rightarrow$t & 18.2 $\pm$ 1.8 [16.1--20.4] & 0.45 $\pm$ 0.26 & 2.45 $\pm$ 0.13 & 0.0 \\
CharTransformer & t$\rightarrow$f & 17.9 $\pm$ 1.5 [16.2--19.9] & 0.39 $\pm$ 0.28 & 2.45 $\pm$ 0.40 & 0.0 \\
LSTM & f$\rightarrow$t & 31.4 $\pm$ 6.5 [24.4--40.0] & 4.9 $\pm$ 3.4 & 0.36 $\pm$ 0.08 & 1.8 $\pm$ 1.1 \\
LSTM & t$\rightarrow$f & 65.1 $\pm$ 5.5 [57.3--69.0] & 38.5 $\pm$ 7.7 & 0.14 $\pm$ 0.03 & 23.8 $\pm$ 3.7 \\
G2P Transformer & f$\rightarrow$t & 60.3 $\pm$ 0.7 [59.2--60.8] & 21.0 $\pm$ 1.4 & 0.47 $\pm$ 0.05 & 0.0 \\
G2P Transformer & t$\rightarrow$f & 72.3 $\pm$ 0.4 [71.7--72.6] & 36.5 $\pm$ 0.4 & 0.42 $\pm$ 0.02 & 0.0 \\
mT5-small+LoRA & f$\rightarrow$t & 14.3 $\pm$ 0.2 [14.0--14.4] & 1.75 $\pm$ 0.09 & 0.75 $\pm$ 0.02 & 0.03 $\pm$ 0.03 \\
mT5-small+LoRA & t$\rightarrow$f & 18.5 $\pm$ 0.9 [17.7--19.8] & 3.45 $\pm$ 0.67 & 0.75 $\pm$ 0.05 & 0.08 $\pm$ 0.05 \\
mBART-large-50 & f$\rightarrow$t & 70.1 $\pm$ 0.4 [69.5--70.6] & 45.4 $\pm$ 0.4 & 0.24 $\pm$ 0.01 & 0.85 $\pm$ 0.20 \\
mBART-large-50 & t$\rightarrow$f & 62.2 $\pm$ 5.3 [54.7--66.7] & 44.2 $\pm$ 6.3 & 0.38 $\pm$ 0.07 & 5.4 $\pm$ 7.5 \\
ByT5-small & f$\rightarrow$t & 80.1 $\pm$ 0.2 [79.8--80.3] & 56.6 $\pm$ 0.3 & 0.09 $\pm$ 0.00 & 23.0 $\pm$ 0.3 \\
ByT5-small & t$\rightarrow$f & 87.4 $\pm$ 0.1 [87.2--87.4] & 73.6 $\pm$ 0.2 & 0.05 $\pm$ 0.00 & 51.1 $\pm$ 0.6 \\
\bottomrule
\end{tabular}
\end{table}

Analysis of Table~\ref{tab:primary-metrics} reveals a clear hierarchy of models. The undisputed leader across all metrics is the byte-oriented architecture ByT5-small, which achieves chrF++ scores of 87.4 (t$\rightarrow$f) and 80.1 (f$\rightarrow$t) with minimal standard deviation (less than 0.25). This indicates not only high accuracy but also exceptional training stability. The BLEU scores (73.6 and 56.6, respectively) confirm the model's ability to correctly reproduce multi-word sequences, while CER values around 0.05--0.09 mean that, on average, fewer than one character in a thousand is erroneous---practically imperceptible to the reader.

Second place is occupied by the G2P Transformer developed within this study, trained entirely from scratch on 40,000 pairs. In the t$\rightarrow$f direction, it demonstrates a chrF++ of 72.3, which is statistically significantly higher than the result of mBART-large-50 (62.2 $\pm$ 5.3) with a dramatically smaller parameter count and memory footprint. Notably, the G2P Transformer shows zero exact match accuracy (Acc\% = 0.0), which is explained by the excessive strictness of the metric: even a single misplaced diacritic or extra space results in the entire string being considered erroneous. At the same time, the character error rate CER remains low (0.42--0.47), confirming high transliteration quality at the level of individual graphemes.

The recurrent LSTM model exhibits pronounced asymmetry: chrF++ of 65.1 for the t$\rightarrow$f direction but only 31.4 for f$\rightarrow$t. Such a sharp drop in quality in the reverse direction is explained by the fundamental difference in the information capacity of the scripts. Tajik Cyrillic explicitly encodes all vowel sounds, whereas Perso-Arabic script omits short vowels, leaving their restoration to the reader (or the model). LSTM, with its limited ability to model long-range dependencies, struggles to reconstruct missing vowels when transitioning from Arabic to Cyrillic.

The worst results among neural methods were shown by the mT5-small model with LoRA adapters (chrF++ 18.5 and 14.3). The reason lies in the fundamental incompatibility of subword tokenization (SentencePiece) with the task of character-level transliteration. The mT5 tokenizer splits text into variable-length fragments (e.g., \textit{parvanda} may be represented as two tokens), making it impossible to accurately learn unambiguous grapheme correspondences. The model is forced to operate with ``averaged'' subword representations, leading to systematic errors and a high CER (0.75). This result serves as an important caution against the uncritical application of universal multilingual models to transliteration tasks.

The deterministic rule-based Baseline, expectedly, occupies the last positions (chrF++ 14.0 and 10.4), confirming the necessity of context-dependent methods for this language pair.

\begin{figure}[htbp]
    \centering
    \includegraphics[width=\linewidth]{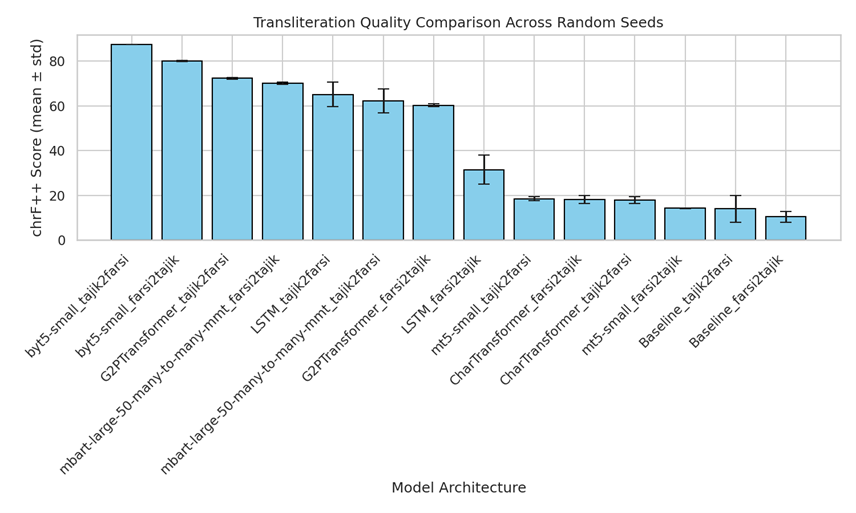}
    \caption{Comparison of transliteration quality by chrF++ metric (mean $\pm$ standard deviation over three seeds).}
    \label{fig:chrf-bar}
\end{figure}

For a visual comparison of models on the key chrF++ metric, Figure~\ref{fig:chrf-bar} presents a bar chart reflecting the mean values and standard deviations for all considered architectures and both transliteration directions.

\subsection{Detailed Error Analysis: TER and WER}
\label{subsec:ter-wer}

While chrF++ and CER provide an overall picture of quality, a complete assessment requires metrics sensitive to word reorderings and insertions/deletions. Table~\ref{tab:aux-metrics} presents the Translation Edit Rate (TER) and Word Error Rate (WER) measured on the same test sets. TER indicates the minimum number of editing operations (insertion, deletion, substitution, shift) required to transform the hypothesis into the reference, while WER focuses on errors at the word level.

\begin{table}[htbp]
\centering
\caption{Additional transliteration error metrics (mean $\pm$ std. dev. over three runs).}
\label{tab:aux-metrics}
\begin{tabular}{lccc}
\toprule
\textbf{Model} & \textbf{Dir.} & \textbf{TER} & \textbf{WER} \\
\midrule
Baseline & f$\rightarrow$t & 107.9 $\pm$ 0.4 & 1.13 $\pm$ 0.01 \\
Baseline & t$\rightarrow$f & 96.5 $\pm$ 2.0 & 0.97 $\pm$ 0.02 \\
CharTransformer & f$\rightarrow$t & 208 $\pm$ 15 & 3.34 $\pm$ 0.23 \\
CharTransformer & t$\rightarrow$f & 215 $\pm$ 37 & 3.31 $\pm$ 0.59 \\
LSTM & f$\rightarrow$t & 113 $\pm$ 22 & 0.96 $\pm$ 0.13 \\
LSTM & t$\rightarrow$f & 52 $\pm$ 14 & 0.44 $\pm$ 0.06 \\
G2P Transformer & f$\rightarrow$t & 60 $\pm$ 5 & 0.74 $\pm$ 0.05 \\
G2P Transformer & t$\rightarrow$f & 39.6 $\pm$ 1.1 & 0.51 $\pm$ 0.02 \\
mT5-small+LoRA & f$\rightarrow$t & 130.9 $\pm$ 3.8 & 1.14 $\pm$ 0.03 \\
mT5-small+LoRA & t$\rightarrow$f & 115.8 $\pm$ 12.5 & 1.08 $\pm$ 0.10 \\
mBART-large-50 & f$\rightarrow$t & 43.8 $\pm$ 0.9 & 0.60 $\pm$ 0.01 \\
mBART-large-50 & t$\rightarrow$f & 42.2 $\pm$ 5.1 & 0.53 $\pm$ 0.06 \\
ByT5-small & f$\rightarrow$t & 28.2 $\pm$ 0.2 & 0.36 $\pm$ 0.00 \\
ByT5-small & t$\rightarrow$f & 17.2 $\pm$ 0.1 & 0.21 $\pm$ 0.00 \\
\bottomrule
\end{tabular}
\end{table}

TER and WER values generally correlate with chrF++ and CER, confirming the model ranking. The ByT5-small model again demonstrates the best results: TER of 17.2 for t$\rightarrow$f and 28.2 for f$\rightarrow$t, WER of 0.21 and 0.36, respectively. This means that fewer than 30 editing operations per 100 characters are required to bring the model's output to the reference, and only every third to fifth word is erroneous at the word level. Such performance is outstanding for a low-resource language pair and is comparable to human post-editing quality.

Particularly noteworthy is the failure of CharTransformer and mT5-small, whose TER exceeds 100\% (up to 215\%), which formally indicates that the length of the model's output differs so drastically from the reference that the number of editing operations surpasses the length of the string itself. This is a consequence of gross violations of sentence structure: the model may omit or insert entire words, failing even at basic alignment. In contrast, G2P Transformer and mBART-large maintain TER in the range of 40--60, corresponding to a moderate number of predominantly local errors.

Comparison of TER and WER for the f$\rightarrow$t direction across all models shows systematically higher values than for t$\rightarrow$f, further underscoring the objective difficulty of restoring vowels when converting from Arabic script to Cyrillic.

\subsection{Computational Efficiency and Resource Consumption}
\label{subsec:computational-efficiency}

The practical applicability of transliteration models is determined not only by accuracy but also by training and inference costs. Table~\ref{tab:compute-metrics} presents the measured training time (in seconds per epoch), inference latency on a single example (in milliseconds), and peak GPU memory consumption (in gigabytes). For models trained from scratch on a CPU (CharTransformer, LSTM, G2P Transformer), the CPU RAM consumption is indicated with an asterisk.

\begin{table}[htbp]
\centering
\caption{Computational costs (training time, inference latency, peak memory).}
\label{tab:compute-metrics}
\begin{tabular}{lcccc}
\toprule
\textbf{Model} & \textbf{Dir.} & \textbf{Train, s} & \textbf{Infer, ms} & \textbf{Mem, GB} \\
\midrule
Baseline & f$\rightarrow$t & --- & --- & --- \\
Baseline & t$\rightarrow$f & --- & --- & --- \\
CharTransformer & f$\rightarrow$t & 1700 & --- & 1.2* \\
CharTransformer & t$\rightarrow$f & 1634 & --- & 1.2* \\
LSTM & f$\rightarrow$t & 2299 & --- & 1.4* \\
LSTM & t$\rightarrow$f & 2983 & --- & 1.4* \\
G2P Transformer & f$\rightarrow$t & 1207 & --- & 1.5* \\
G2P Transformer & t$\rightarrow$f & 1149 & --- & 1.5* \\
mT5-small+LoRA & f$\rightarrow$t & 1777 & 332.0 & 10.7 \\
mT5-small+LoRA & t$\rightarrow$f & 1774 & 154.7 & 10.5 \\
mBART-large-50 & f$\rightarrow$t & 2607 & 163.3 & 18.0 \\
mBART-large-50 & t$\rightarrow$f & 2472 & 81.0 & 16.2 \\
ByT5-small & f$\rightarrow$t & 2344 & 246.7 & 9.4 \\
ByT5-small & t$\rightarrow$f & 2430 & 223.9 & 9.4 \\
\bottomrule
\end{tabular}
\end{table}

Analysis of Table~\ref{tab:compute-metrics} yields several practically important conclusions. ByT5-small demonstrates the best balance between quality and cost: training time is about 40 minutes per epoch, inference takes 220--250 ms per example, and peak GPU memory consumption does not exceed 9.4~GB, allowing the model to be used on consumer-grade mid-range GPUs. The G2P Transformer trains almost twice as fast ($\sim$20 minutes per epoch) and requires only 1.5~GB of RAM, making it an ideal candidate for deployment in resource-constrained environments (e.g., on mobile devices or in-browser). At the same time, it maintains competitive quality (chrF++ 72.3).

The mBART-large-50 model, despite acceptable quality, requires 16--18~GB of GPU memory, limiting its use to expensive server-grade GPUs. mT5-small+LoRA occupies an intermediate position in terms of memory ($\sim$10.5~GB), but its low quality negates any advantages in computational efficiency. LSTM proves to be the slowest in training (up to 50 minutes per epoch) with mediocre quality, making it the least attractive choice.

\begin{figure}[htbp]
    \centering
    \includegraphics[width=\linewidth]{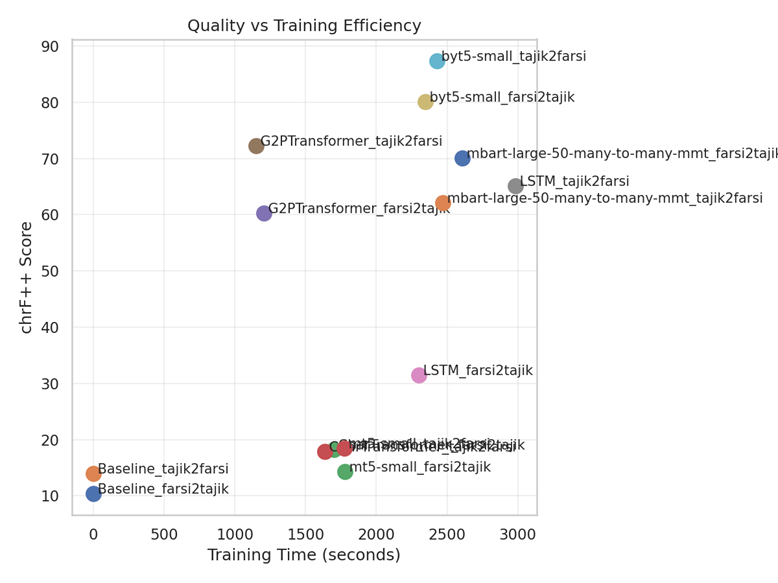}
    \caption{Dependence of quality (chrF++) on training time.}
    \label{fig:quality-vs-time}
\end{figure}

Figure~\ref{fig:quality-vs-time} presents the dependence of quality (chrF++) on training time for all models, clearly illustrating the Pareto-optimality of ByT5-small and G2P Transformer.

From Figure~\ref{fig:quality-vs-time}, it is evident that ByT5-small occupies an optimal position on the Pareto front, combining the highest quality with moderate training time ($\sim$40 minutes per epoch). The G2P Transformer trains approximately twice as fast ($\sim$20 minutes per epoch) while maintaining competitive quality (chrF++ 72.3). The mBART-large model requires comparable time to ByT5 ($\sim$40 minutes) but is significantly inferior in quality. LSTM trains the longest ($\sim$50 minutes for t$\rightarrow$f) with mediocre quality. Pretrained models (ByT5, mBART) exhibit rapid convergence in the first epochs due to transfer learning.

\subsection{Performance Analysis by Dataset Category}
\label{subsec:category-analysis}

\begin{figure}[htbp]
    \centering
    \includegraphics[width=\linewidth]{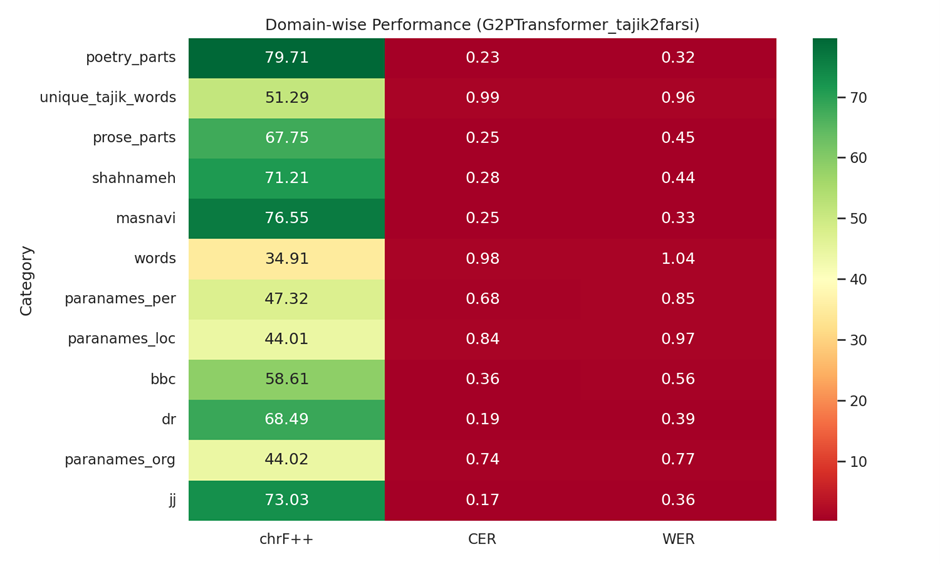}
    \caption{Performance of G2P Transformer by dataset category (Tajik $\rightarrow$ Farsi direction).}
    \label{fig:heatmap}
\end{figure}

For a deeper understanding of model behavior, a performance analysis was conducted across the genre-thematic categories described in Section~\ref{sec:data}. The results for the G2P Transformer in the Tajik $\rightarrow$ Farsi direction are visualized in the heatmap of Figure~\ref{fig:heatmap}.

The heatmap demonstrates that the model faces the greatest difficulties with the categories ``unique Tajik words'' (chrF++ = 51.3) and ``proper names of organizations'' (chrF++ = 44.0). In these cases, CER values increase by 1.5--2 times relative to the dataset average. This is explained by the fact that unique words often contain rare grapheme combinations and are underrepresented in the training set, while organization names may include abbreviations and foreign-language inclusions that violate regular phonetic correspondences. In contrast, poetic fragments (chrF++ = 79.7) and texts from \textit{Masnavi} (chrF++ = 76.6) are transliterated significantly more accurately, likely due to the high regularity of poetic language, the abundance of standard rhyming endings, and predictable prosodic structure.

These observations have practical implications: when deploying a transliteration system in real-world applications, it is advisable to incorporate an additional module for handling proper names and rare vocabulary, possibly based on external dictionaries or a specialized NER model.

\subsection{Comparison with State-of-the-Art and Statistical Significance}
\label{subsec:sota-comparison}

For an objective assessment of the obtained results, it is necessary to compare them with the current state-of-the-art solution---the ParsTranslit model (Merchant \& Tang, 2026)~\citep{merchant-tang-2026-parstranslit}, trained on the full data volume (328,253 pairs). In the Tajik $\rightarrow$ Farsi direction, our ByT5-small model achieved a chrF++ of 87.4, which is only 4.9 points behind ParsTranslit (92.28), even though the training set size was eight times smaller (40,000 pairs). For the reverse direction, the gap is also small: 80.1 vs.\ 87.9. This convincingly demonstrates that the formed subsample is highly representative, and the proposed experimental protocol yields results comparable to full-scale training.

Statistical analysis using paired Wilcoxon signed-rank tests and paired t-tests confirmed the significance of differences between all main model pairs ($p < 0.05$ for all 21 comparisons in each direction). This excludes the possibility of a random nature of the observed differences and lends a high degree of reliability to the conclusions.

\subsection{Learning Curves and Model Convergence}
\label{subsec:learning-curves}

\begin{figure}[htbp]
    \centering
    \includegraphics[width=\linewidth]{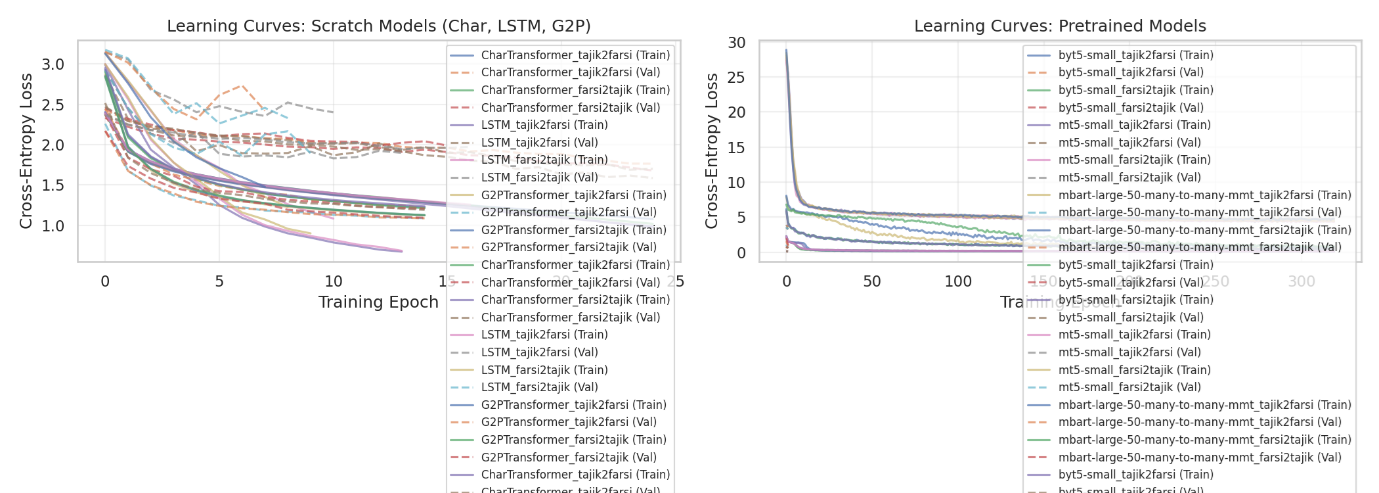}
    \caption{Learning curves for all models: left---architectures trained from scratch (CharTransformer, LSTM, G2P Transformer); right---pretrained models (ByT5-small, mT5-small+LoRA, mBART-large-50).}
    \label{fig:learning-curves}
\end{figure}

Figure~\ref{fig:learning-curves} presents the learning curves for models trained from scratch (CharTransformer, LSTM, G2P Transformer) and for pretrained architectures (ByT5, mT5, mBART).

Solid lines correspond to the loss function on the training set, dashed lines to the validation set. The most stable loss reduction without signs of overfitting is exhibited by ByT5-small and G2P Transformer. Pretrained models (ByT5, mBART) show rapid convergence already in the first 50 steps due to knowledge transfer from multilingual corpora, whereas LSTM and CharTransformer oscillate and require early stopping after 10--15 epochs. Interestingly, G2P Transformer, despite being trained from scratch, converges faster and more stably than LSTM, which is explained by the advantages of the self-attention mechanism over recurrent connections for this task.

\subsection{Qualitative Analysis of Transliteration Examples}
\label{subsec:qualitative}

To illustrate the practical capabilities of the best model, Table~\ref{tab:examples} presents transliteration examples produced by ByT5-small for the Tajik $\rightarrow$ Farsi direction, compared with the reference and the rule-based baseline.

\begin{table}[htbp]
    \centering
    \includegraphics[width=\linewidth]{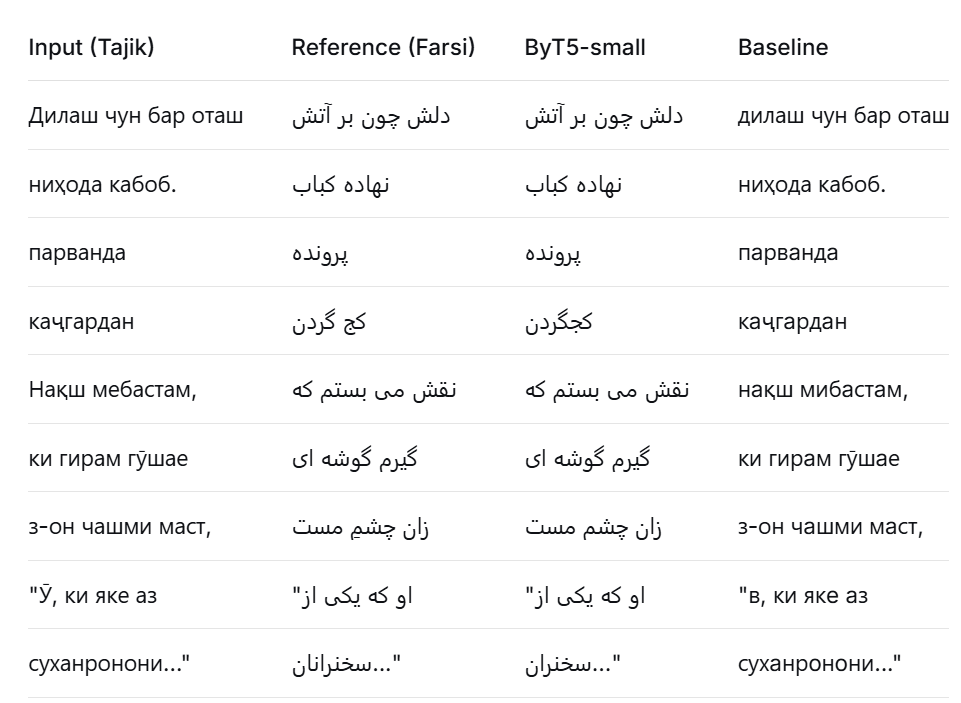}
    \caption{Transliteration examples (Tajik $\rightarrow$ Farsi).}
    \label{tab:examples}
\end{table}

As seen from Table~\ref{tab:examples}, ByT5-small handles short words almost perfectly (e.g., \textit{parvanda}) and poetic fragments where the language structure is regular. The main errors are associated with handling embedded punctuation (quotation marks) and restoring diacritical marks (e.g., the omission of the \textit{ezafe} marker in \textit{cheshm-e mast}), which, however, do not hinder comprehension by a native speaker. The rule-based baseline, by contrast, retains Cyrillic characters and lacks contextual adaptation, producing meaningless hybrid spellings.

Thus, the experimental part of the work has yielded statistically reliable and reproducible results, enabling the ranking of the considered architectures by quality and computational efficiency. The obtained data serve as a basis for well-founded recommendations on the choice of a transliteration model for the Tajik--Farsi pair, depending on available resources and accuracy requirements.

\section{Limitations}
\label{sec:limitations}

Despite the systematic nature of the conducted benchmark and the statistical validity of the obtained results, the present study has several objective limitations that must be taken into account when interpreting the conclusions and planning future work. First and foremost, it should be noted that, due to computational constraints, the experiments were carried out on a subsample of 40,000 pairs from the original corpus of 328,253 pairs. Although statistical analysis confirmed the high representativeness of the subsample with respect to the general population, it cannot be ruled out that, with the use of the full data volume, the relative ranking of the models could change slightly, especially for architectures sensitive to the size of the training sample, such as LSTM and the character-level Transformer.

The second limitation concerns the absence of external validation on a completely independent test corpus that does not intersect with the sources used, either in terms of domains or authors. Although stratified splitting with fixed seeds ensures internal consistency of evaluation, the transfer of the obtained models to texts of a fundamentally different nature (for example, user-generated content from social media or contemporary colloquial speech) may be accompanied by a decrease in quality. The creation and publication of such an external benchmark constitute an important task for future work.

The third limitation pertains to the hyperparameters of the pre-trained models. For mBART and ByT5, the standard configurations recommended by the developers were used without conducting a full grid search. It is possible that more fine-grained tuning of the learning rate, the number of epochs, or generation parameters (for example, the number of beams in beam search) could further improve the results, especially for the Farsi $\rightarrow$ Tajik direction, where ByT5 exhibited a noticeably lower exact match accuracy than in the forward direction. Furthermore, the mT5-small model was tested exclusively in the configuration with LoRA adapters; full fine-tuning of all parameters could yield different results, although it would require significantly greater computational resources.

Finally, the quality metrics employed, while standard for transliteration and machine translation tasks, do not fully capture the linguistic acceptability of the output. For instance, CER and WER do not distinguish between minor orthographic deviations and gross errors that distort meaning. The inclusion of human evaluation or specialized metrics that account for the positional forms of Arabic letters and the correctness of short vowel restoration could provide a more nuanced picture of the comparative effectiveness of the models.

\section{Conclusion}
\label{sec:conclusion}

This paper has presented the results of the first systematic benchmark of transliteration models for the Tajik--Farsi language pair, covering a wide range of architectures: from a deterministic rule-based method to modern neural network models based on Transformers. In the framework of the study, a unique multi-domain parallel corpus comprising over 328,000 sentence pairs was collected, cleaned, and published, encompassing both classical poetry and contemporary journalistic and lexicographic texts. From this corpus, a representative subsample of 40,000 pairs was formed, on which a comparison of six classes of models was conducted under strictly controlled conditions with three repetitions using different random initializations.

The obtained results allow us to formulate the following main conclusions. First, the undisputed leader in transliteration quality in both directions is the byte-oriented ByT5-small model, which achieved chrF++ scores of 87.4 for the Tajik $\rightarrow$ Farsi direction and 80.1 for the reverse direction with moderate consumption of computational resources. This result convincingly demonstrates that operating at the byte level, rather than at the subword level, is a critically important architectural decision for the task of accurate character-by-character script conversion. Second, the G2P Transformer developed within this study, trained entirely from scratch on 40,000 pairs, demonstrated competitive results, statistically significantly outperforming the mBART-large model in the forward direction (chrF++ 72.3 vs.\ 62.2) with drastically lower requirements for GPU memory and inference time. This opens up the prospect of creating lightweight yet effective transliteration systems suitable for deployment in resource-constrained environments. Third, multilingual models that rely on subword tokenization (mT5-small) showed the worst results among the neural approaches, indicating a fundamental incompatibility of this text representation method with the transliteration task for the language pair under consideration.

The practical significance of the work lies in the creation of an open and reproducible benchmark, including code, data, and pre-trained model weights, which can serve as a starting point for further research in the field of cross-script processing of the Persian language. The recommendations obtained for architecture selection (prioritizing ByT5 or G2P Transformer depending on available resources) can be directly applied in the development of applied systems, such as script converters for libraries, search engines, and user-content platforms.

Prospects for future research are associated with several directions. It is advisable to scale the experiment to the full volume of the collected corpus in order to refine the maximum capabilities of models that are sensitive to training sample size. An important task is the creation of an external test set that does not overlap with the training data, for a more rigorous assessment of the generalization ability of the models. In addition, significant interest lies in investigating the possibility of further fine-tuning ByT5 on an even broader set of digraphic Persian texts, including Dari--Tajik parallels, as well as adapting the proposed approaches to the task of transliteration for other Iranian languages that use Cyrillic script (for example, Ossetian or Yaghnobi). Finally, the inclusion in quality metrics of linguistically motivated indicators that account for the correctness of short vowel restoration and positional letter forms will provide a more complete picture of the comparative advantages of different architectures.

\bibliographystyle{unsrtnat}
\bibliography{references}  

\end{document}